# LEARNING CHESS WITH LANGUAGE MODELS AND TRANSFORMERS

Michael DeLeo and Erhan Guven

Whiting School of Engineering, Johns Hopkins University, Baltimore, USA

***ABSTRACT***

*Representing a board game and its positions by text-based notation enables the possibility of NLP applications. Language models, can help gain insight into a variety of interesting problems such as unsupervised learning rules of a game, detecting player behavior patterns, player attribution, and ultimately learning the game to beat state of the art. In this study, we applied BERT models, first to the simple Nim game to analyze its performance in the presence of noise in a setup of a few-shot learning architecture. We analyzed the model performance via three virtual players, namely Nim Guru, Random player, and Q-learner. In the second part, we applied the game learning language model to the chess game, and a large set of grandmaster games with exhaustive encyclopedia openings. Finally, we have shown that model practically learns the rules of the chess game and can survive games against Stockfish at a category-A rating level.*

***KEYWORDS***

*Natural Language Processing, Chess, BERT, Sequence Learning.*

## 1. INTRODUCTION

One of the oldest board games, chess is also one of the most researched computational problems in artificial intelligence. The number of combinational positions is around 10^50 according to [1] and this makes the problem ultimately very challenging for even today's computational resources. State of the art solution to learning the chess game by a computer has two parts, generating valid board positions and evaluating their advantage to win the game. Like an optimization problem, generating possible and promising positions is analogous to a feasible optimization surface and is built by a tree data structure representing each position reached from a previous position. Evaluating a position involves the chess game knowledge, such as how a piece moves, their values, the position of the king, opponent piece positions, and existence of a combination of moves that can lead to a forced mate, are among numerous calculations to find the winning move or a combination of moves.

Stockfish [2], one of the best chess engines today, uses the minimax search with alpha-beta pruning improvement [3] by avoiding variations that will never be reached in optimal play. The computationally infeasible search is avoided where it is possible to infer the outcome of the game, such as in a deterministic mating attack discovered by the search tree. IBM Deep Blue chess computer [4] used dedicated processors conducting tree searches faster on hardware to beat the world champion Garry Kasparov who is considered as one of the best human chess players ever. Alpha Zero [5] uses plain randomly played games without any chess knowledge but learns the moves from the game. A general-purpose reinforcement learning algorithm, and a general-purpose tree search algorithm are used to conduct combination of moves. The deep learning engine learns the game as much as the hand-crafted knowledge injected in the Stockfish evaluation engine.

In this study, we evaluated and analyzed a second approach to learn chess, specifically using BERT transformer to extract the language model of chess by observing the moves between players. Language models extracted by the state-of-the-art models such as GPT-3 [6] are considered as few-shot learners. Among many other statistical information, a language model can involve rare patterns represented by only a few counts in a histogram which can be extracted by BERT transformers.

Our study evaluates the BERT language model starting from a simple game, then towards increasing the complexity of the game as in chess. First, we analyzed the application of a language model to a simpler game Nim due to its smaller size allowing a complete analysis. We conducted experiments of Nim games between a Guru player which knows winning solutions to Nim as a rule-based approach versus a random player blindly making valid Nim moves and evaluate the performance of the language model. Next, we analyzed the model between a random, a Guru, and a Q-learner player with a controlled number of random games. And finally, we applied the model to the chess game by grandmaster games covering all possible openings from the chess opening encyclopedia [7].

A literature survey shows only a few very recent papers [8, 9] have applied NLP methods to chess but none of them used board/move text based on a grammar pattern to encode the game. As a novelty, the method in this paper encodes the game positions and moves in a specific text pattern based on Forsyth-Edwards Notation [10] which is possibly easier to be learned than a full game in Portable Game Notation format [10]. Starting from the opening position, PGN conveys a position virtually between the moves without explicitly encoding. In a board game each position and move pair can be thought of a sentence passed to the other party. Thus, these sentences are learned by the language model, and they are somewhat order independent. The following sections will describe the NLP method and analyze its performance towards learning board games that use text representation of each position and move.

## 1.1. Chess State of the Art

Stockfish is under the GPL license, open source, and still one of the best chess programs making it a suitable candidate to teach a natural language model. There are two main generations of Stockfish which are Classical and NNUE. The latter is stronger of the two, and for the purposes of this research is what will be focused on. In this version, Stockfish relies on a neural network for evaluation rather than its previous method of relying on a search tree. NNUE stands for Efficiently Updateable Neural-Network [11]. This network is a lightweight fully connected neural network that gets marginally updated depending on the state space of the board, which is an optimization technique to improve its performance.

Alpha Zero is a deep reinforcement learning algorithm with a Monte Carlo Tree Search (MCTS) algorithm as its tree search algorithm [8]. MCTS is a probabilistic algorithm that runs Monte Carlo simulations of the current state space to find different scenarios [3]. An example of the MCTS being used for a game of tic tac toe is shown in Figure 1. Notice the tree branches off for various game choices. This is a critical component of the Alpha Zero model in that it allows it to project/simulate potential future moves and consequences. MCTS was chosen by the deep mind team as opposed to using an Alpha Beta search tree algorithm because it was more lightweight.
Alpha Zero is famous for beating Stockfish in chess with 155 wins out 1000 games. Stockfish won 6 games [8]. There is some debate however as to if more hardware would have helped Stockfish. Nonetheless, the main advantage of Alpha Zero to Stockfish is that it is a deep learning model which can play itself millions of times over to discover how to best play chess. One of the impracticalities of it is that it is not open source however, and proprietary to DeepMind.

## 1.2. Chess Text Notation

In this study the chess notation is based on coordinate algebraic notation. This notation is based on the chess axes where {a, b, ..., h} is the x axis, and {1, 2, ..., 8} is the y axis and represents two coordinates {(x!, y!, ), (x", y")}. The first coordinate set represents the initial position, and the second set represents the position the piece moves to [10]. This notation is chosen to represent move states rather than other notations is because of its uniformity and how many tokens it would take to represent a full game position.

The Forsyth-Edwards Notation (FEN) is a notation that described a particular board state in one line of text with only ASCII characters [10]. A FEN sequence can completely describe any chess game state while considering any special moves. We use this notation to describe our board state in our experiments. An example of the FEN sequence is shown in Figure 1, this record represents the initial chess state.

```
rnbqkbnr/pppppppp/8/8/8/8/PPPPPPPP/RNBQKBNR w KQkq - 0 1
```

Figure 1. Example FEN Sequence

1. Piece placement: each piece is represented by a letter r, n, b, q, etc. and the case indicates the player where uppercase is white, and lowercase is black. The rank of each section of piece is described by Standard Algebraic Notation (SAN) [10], and this describes the positions of those pieces.
2. Active color: represented by a “w” meaning white’s turn is next, and “b” meaning black’s turn is next.
3. Castling Availability: There are five tokens to represent castling availability, “-“ no one can castle, “K” white can castle king side, “Q” white can castle queen side, “k” black can castle king side, and “q” black can castle queen side.
4. En Passant.
5. Half Move Clock: Starts at zero and represents the number of moves since the last capture or pawn advance.
6. Full Move Clock: Starts at 1, increments after black’s move [10].

FEN is particularly useful because it provides a complete stateless representation of the game state in a single character sequence. This is possible because chess is a game where there are no unknowns, and everything represented visually is everything there is to the game space. These are the reasons for why we chose the game of chess and chose these notations for our experimentations.

## 1.3. Nim Game

Nim is a game of strategy between two players in which players remove items from three piles on each turn. Every turn the player must remove at least one item from exactly one of the piles. There are two different versions of the game goal: the player who clears the last pile wins or the player who has to take the last piece loses the game.

### 1.4. BERT Model

The BERT model (Bidirectional Encodings for Representations of Transformers) is a language model which is designed to pretrain on bidirectional representations on unlabeled text by jointly conditioning on context from both the right and left sides [12]

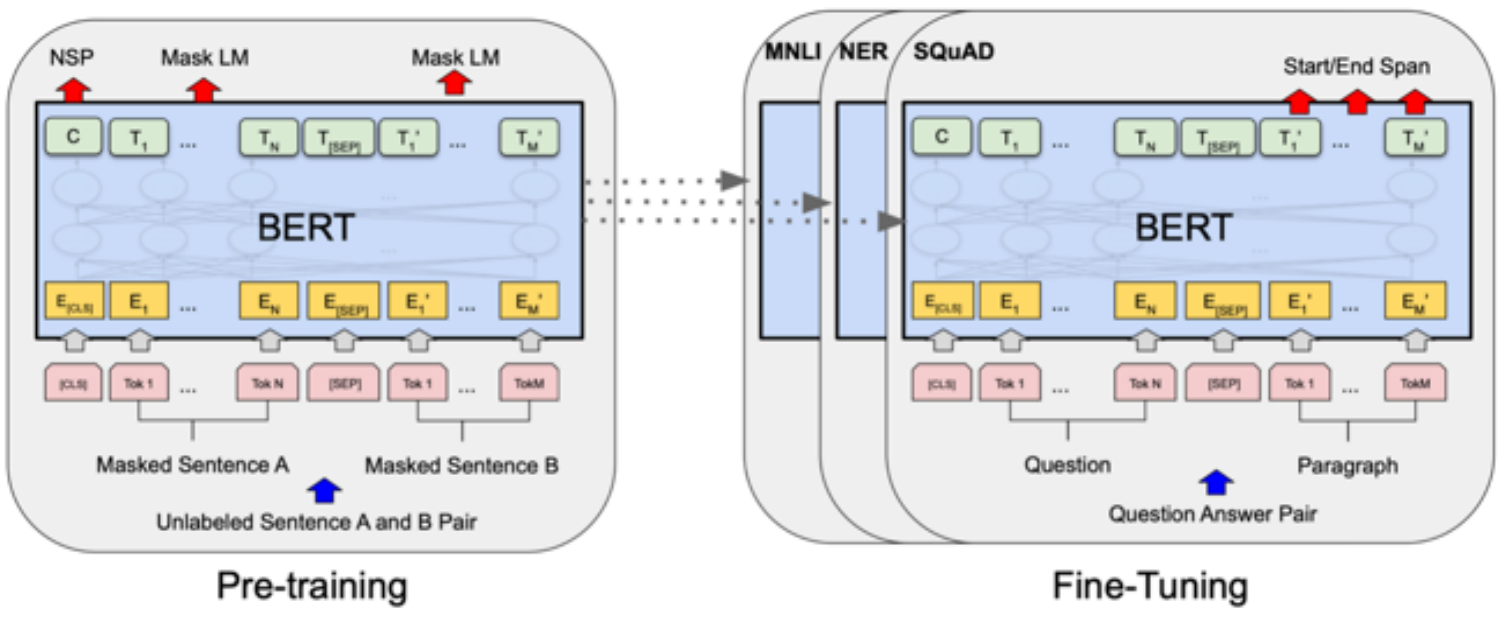

Figure 2. BERT Architecture [12]

The Bidirectional Encoder Representations from Transformers (BERT) model is a supervised model, that achieved state of the art on Q&A tasks before GPT. It's a lightweight, deep learning model that is trained to learn bidirectional representations of context in unlabeled text. The general architecture can be seen in Figure 2, and it should be noted that it is like GPT-1 in terms of its architecture and size [13].

## 2. Methodology

The objective of our study is to train a transformer model on text sequence datasets in such a way that it can learn to accurately play and understand the games. We apply the BERT model to both Nim and Chess. In this section we will lay our procedure for procuring the data for these experiments as well as our methodology for training the transformer.

### 2.1. Nim Data Collection

Our Nim experiment consisted of using three agents: a random player, a guru player, and a Q-learner. Each experiment is initialized to three piles, and ten items per pile (i.e. [10,10,10]). Equal number of games are played between each player taking equal number of times as the first player. The version of the Nim game in this study favors the first player. As a result, when two Guru players play against each other the first player wins roughly 95% of the time.

A variety of games positions are created by randomly creating the pile positions starting from [1,1,1] to [10,10,10] so that occasionally the random player can also win against a Guru just because the starting position was a lucky one. These played and stored games are used to train a Q-learner and a BERT learner later on in the experimentation.

Additionally, when Nim games are collected the pile positions of the game are randomized throughout each recorded game. This was to increase the level of difficulty for a model to learn from the sequences.

## 2.2. Chess Data Collection

The chess experiment uses Stockfish 14, and python 3.9. The Stockfish engine is configured to use NNUE, with one thread, default depth, and one for the value of MultiPV. The max depth that can be set is 20, however that slows the experiment down too much and so a value of 1 is chosen for the sake of getting a large dataset. Additionally, Stock- fish is set to an ELO rating of 3900. To gather a large amount of data, one million games of chess are played. A timeout for moves is set for 200 to discourage runaway stalemate games. The data collection activity takes about 4-6 days.

```
rnbqkbnr/pppppppp/8/8/8/8/PPPPPPPP/RNBQKBNR w KQkq - 0 1 [MOVESEP] f2f4
```

Figure 3. Example Chess Sequence

1. The basic routine of the program is to initialize a fresh game with the Stockfish engine, and the stated configurations
2. Select the best move from Stockfish and submit for each player until the game is over
3. Record the moves (FEN and algebraic coordinate) as they are selected and store
4. At six moves end the game, delimit each set of moves with the next line tokens and perform post-processing

An example of the data returned by a game of chess is shown in Figure 3 where each line consists of a FEN position, the player, and the next move chosen. This example is the opening chess board, followed by a separator token and a move for F2 to F4.

## 2.3. Pretraining BERT

For each experiment, once the data is generated, the BERT Word Piece Tokenizer is trained on the entire set of data such that it can get a full scope of the sequences. Since information is encoded into words and letters being capitalized, the tokenizer must accommodate for this. Therefore, the vocabulary includes capitalization.

We utilize the datasets hugging face library to load all our datasets and deliminate by the end of line token. Those datasets are tokenized and collated in parallel then split into training and testing sets with a 20% split. For the training procedure, we use the hugging face trainer with a 15% MLM (masked language modelling, refer to [12]) probability.

To provide inference with the model, a hugging face pipeline is used where the state sequence is provided and a [MASK] token is placed at the end where the move token would be. For example, a sequence for Nim would be a10/b10/c10 G – [MASK], where the pipeline would fill in the MASK token for the move.

## 2.4. Initial Analysis of BERT Model on Nim

As a few shot learner and as an unsupervised learner [6, 14] BERT language model can extract patterns that are expressed only a few times and in midst of very high noise. The following experiment used the language model trained using the games between Guru and random players. A number of games are played between the Guru which is a rule based player, and a random player generating random but valid moves. Since the Nim game outcome heavily depends on the first player move (like Tic Tac Toe), an equal number of games are played by swapping the first player. Each game start from a non- zero number of pieces in three piles, so that a Guru player can lose a game against a random player since it might be given a losing position in the first place. The number of possible positions or the feature space size is 113 (equals 1331) for three piles and 10 pieces to start the game. Theoretically one needs at least this many positions to fill up the feature space for a Guru to make a move so the game data would have at least one sample of every board position and winning (or the "right") move by Guru. Note that for the Q-learner, such a learning approach takes close to 300k games (against a random player) to be able to be on par with a Guru player [15].

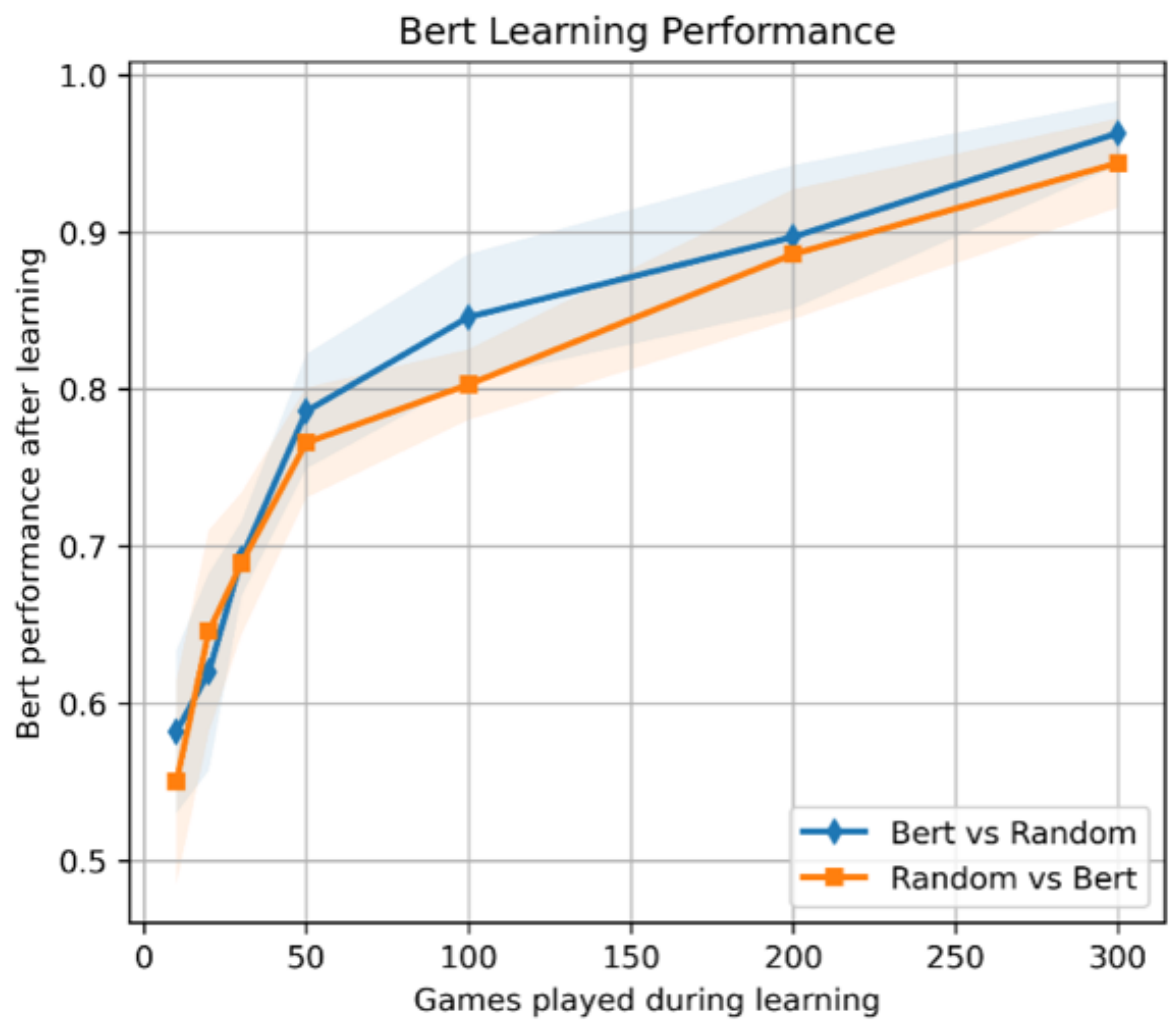

Figure 4. BERT Learning Nim Game from Random Player

The experiment trains a transformer by a certain number of games defined as a match, played between the Guru and the random player. Every training starts from reset and the trained model is used by the BERT player to make a move. An average game against Guru by the random player takes empirically ~6.5 many moves. Thus, 10- game match of Guru-random and an additional 10-game match of random-Guru would provide around 130 unique moves possibly. This space covers only 10% of the feature space (game board) presenting an almost impossible learning problem. Against all odds, as shown by the experimentation, the BERT player learns every move that Guru makes and plays accordingly when faced a random player. The range of number of games (match size) is changed from 10 to 300 where the latter makes the BERT player an excellent challenger for the random player. This is the direct result of the few-shot learning method presented by the transformers.

## 3. EXPERIMENTAL RESULTS

Following the methodology stated of performing collections of data into a standard dataset format, data was collected for both Nim and chess experiments. Some of the characteristics for these datasets are listed in Table 1 - Dataset Metrics.

Table 1. Dataset Metrics

| Methodology Metric | Nim | Chess |
|---|---|---|
| Number of Games | 30,000 | 30,000 |
| Total Unique Game States | 7973 | 2575 |
| Total Unique Moves | 30 | 892 |
| Dataset Length | 423,480 | 2657 |
| Average Sequence Length | 15.09 | 74.28 |
| Dataset Size (MB) | 8.1 | 167.9 |

Notably with our method of data collection and data format, choosing longer sequences to represent a system will cause the dataset's memory size to grow by an order of M where M is the current dataset length. This caused issues when we initially tried to generate a chess dataset that contained one million games and created an 8 GB text corpus. This is also one of the reasons we added the Nim experiment, to test our hypothesis on a smaller scoped dataset.

The hardware used for the experimentation is an RTX-A6000, an i9 processor, and 128 GB of RAM.

## 3.1. Results Nim

Following the data collection, two BERT word piece tokenizers were trained on each variant of the Nim datasets: X/W and Player ID. The vocabularies for each tokenizer were relatively small and are shown in Figure 5 and Figure 6 respectively. The vocabulary size maxed out around 60 tokens for each tokenizer because the game of Nim is not that complicated in sequence form.

{'a3': 43, 'a1': 33, '8': 15, 'b8': 59, 'b9': 61, 'b1': 35, '4': 11, '##9': 23, 'c8': 58, 'G': 17, 'c4': 46, '##5': 31, '##6': 28, 'b3': 44, '##2': 32, 'c5': 48, 'c3': 42, '9': 16, '2': 9, 'c1': 34, 'a': 20, '6': 13, 'a10': 65, 'c0': 38, 'a8': 57, 'R': 19, '7': 14, 'b4': 45, 'c6': 51, 'a7': 56, 'c7': 54, '##3': 27, '[MASK]': 4, 'c9': 60, '[PAD]': 0, 'a5': 50, '##1': 24, '[CLS]': 2, '-': 5, 'b5': 49, 'b10': 64, '##7': 26, 'a9': 62, '1': 8, '0': 7, 'c': 22, 'a0': 36, '3': 10, 'a6': 53, '5': 12, 'Q': 18, 'a4': 47, 'c10': 63, 'b0': 37, '##0': 25, 'a2': 41, '/': 6, '##4': 29, '[UNK]': 1, 'b6': 52, 'c2': 40, 'b2': 39, 'b7': 55, '##8': 30, 'b': 21, '[SEP]': 3}

Figure 5. BERT Tokenizer Tokens for Nim with Player IDs' G, Q and R.

{'a0': 35, '/': 6, '9': 16, 'b1': 34, 'c1': 33, 'a6': 52, '7': 14, 'c9': 59, '##1': 24, '[UNK]': 1, '[CLS]': 2, 'c3': 42, 'b9': 61, 'a9': 60, 'c8': 56, '[MASK]': 4, 'b': 20, '##7': 30, 'c2': 39, 'a2': 40, 'b3': 41, '##8': 22, '1': 8, 'c0': 37, 'a1': 32, 'c4': 44, '-': 5, '##5': 26, 'b8': 57, '2': 9, 'W': 17, '5': 12, 'a10': 63, 'c7': 54, 'b0': 36, 'c10': 64, '6': 13, '##6': 27, '##0': 25, '[PAD]': 0, 'c': 21, '##4': 31, '0': 7, '4': 11, 'b5': 48, '3': 10, 'a4': 46, 'a3': 43, '[SEP]': 3, 'b2': 38, 'X': 18, '##2': 23, 'a': 19, 'b6': 50, 'a7': 55, '8': 15, 'b10': 62, 'c6': 51, '##3': 29, 'a8': 58, 'b4': 45, '##9': 28, 'a5': 49, 'b7': 53, 'c5': 47}

Figure 6. BERT Tokenizer Tokens for Nim with Win States

We can verify the tokenizers captured the game state tokens and move tokens by inspecting their vocabulary. Since for Nim, the game state is being represented by a letter a, b, or c and a quantity we can see that those tokens do exist in the vocabularies.

Recall that two datasets were generated with partitions to designate artificial noise, the first had a special indicator for which agent made this move and the other had an indicator to as if this move won the game. We trained a fresh BERT model on each partition of each dataset and put each model into a roster where each agent played every single other agent, the results are below. The evaluation was performed with 1000 games of every permutation of every agent for each level of randomness for a total of 5000 games. The total wins for each partition were collected and that is what is shown in Figure 7, Figure 8, Figure 9, and Figure 10. For each level of randomness and for each graph, it took 20 minutes to train the BERT model.

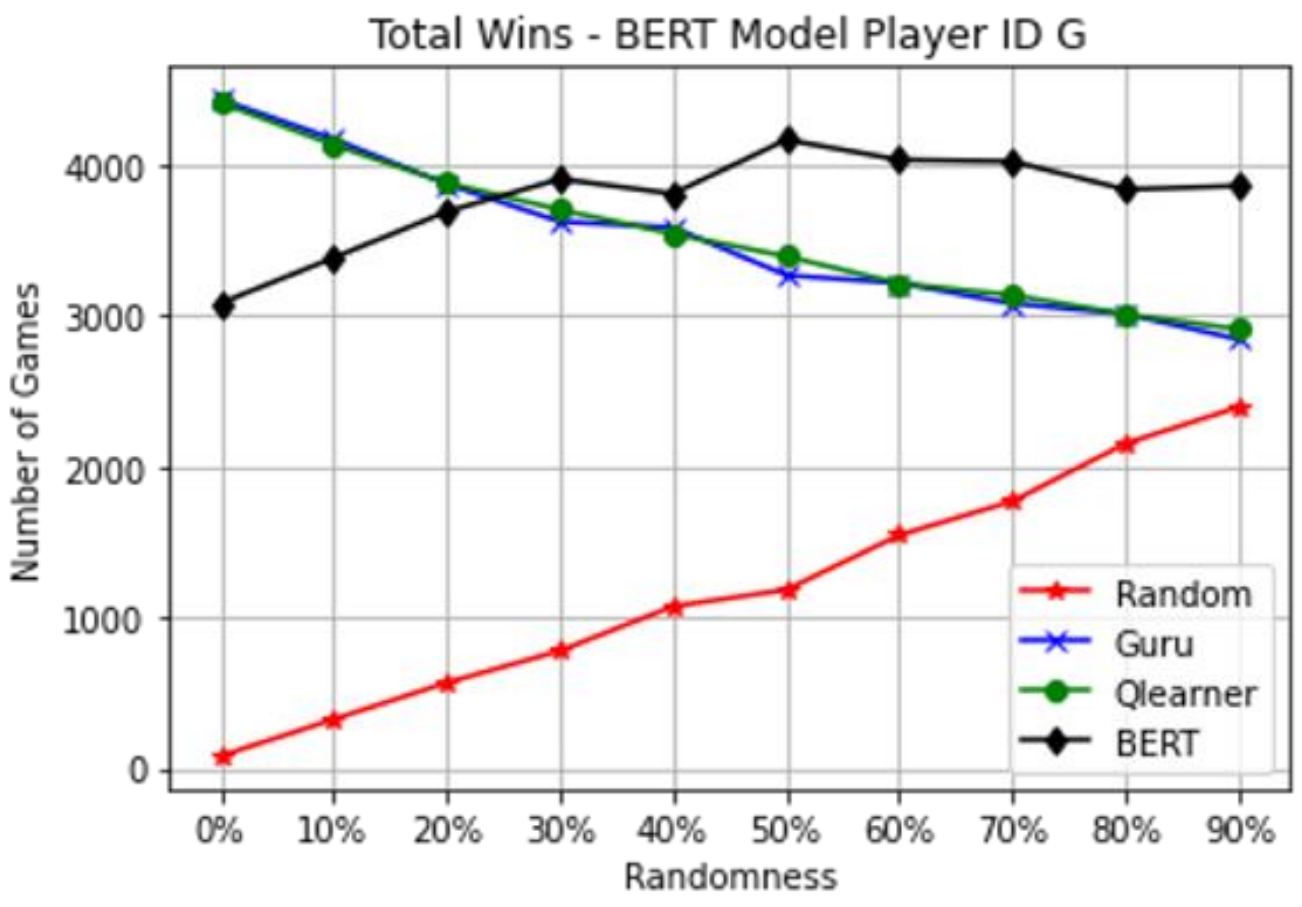

Figure 7. Nim Player ID G. The BERT model inferenced and played with the Guru (G) ID token being specified.

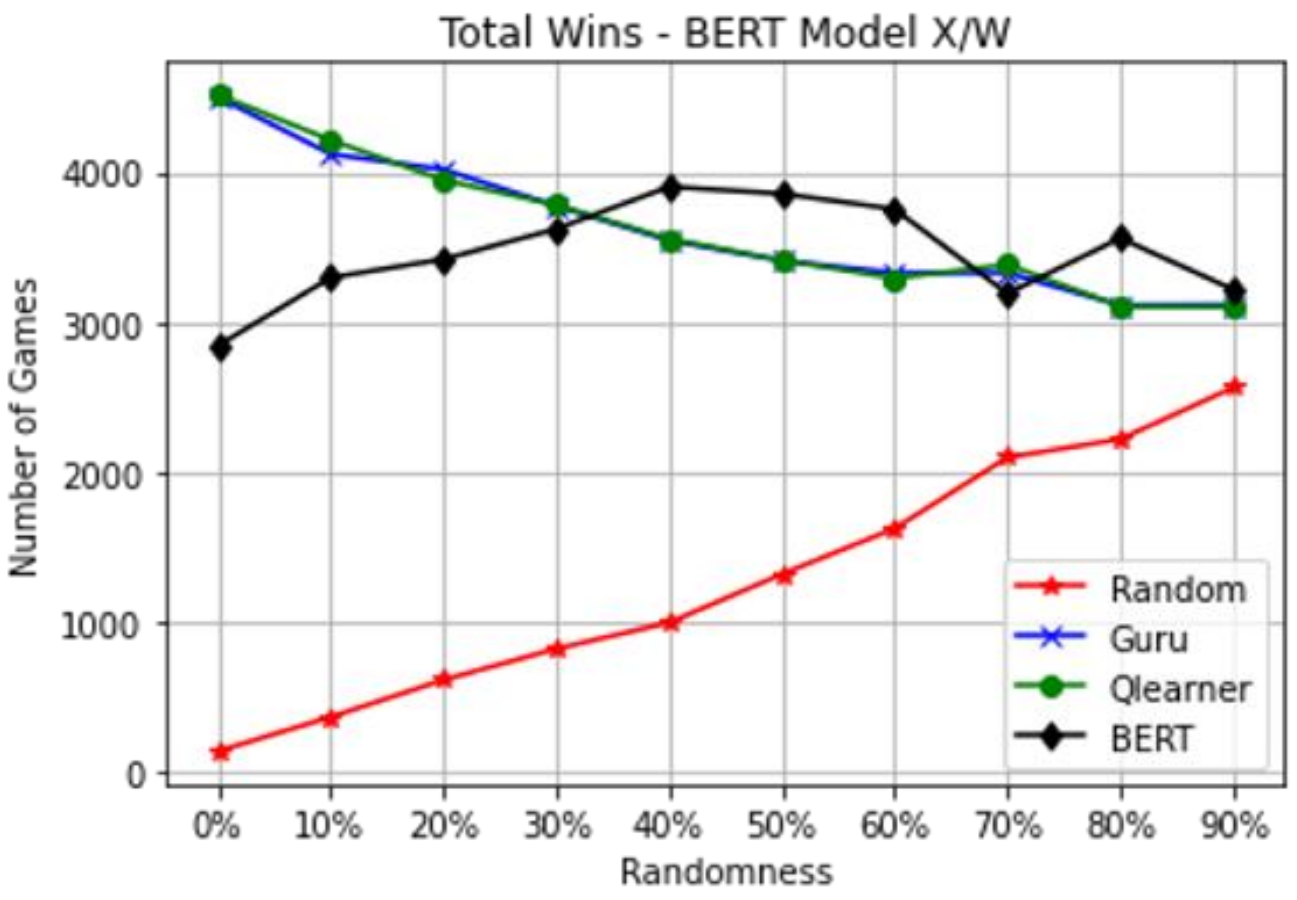

Figure 8. Nim X/W Win State

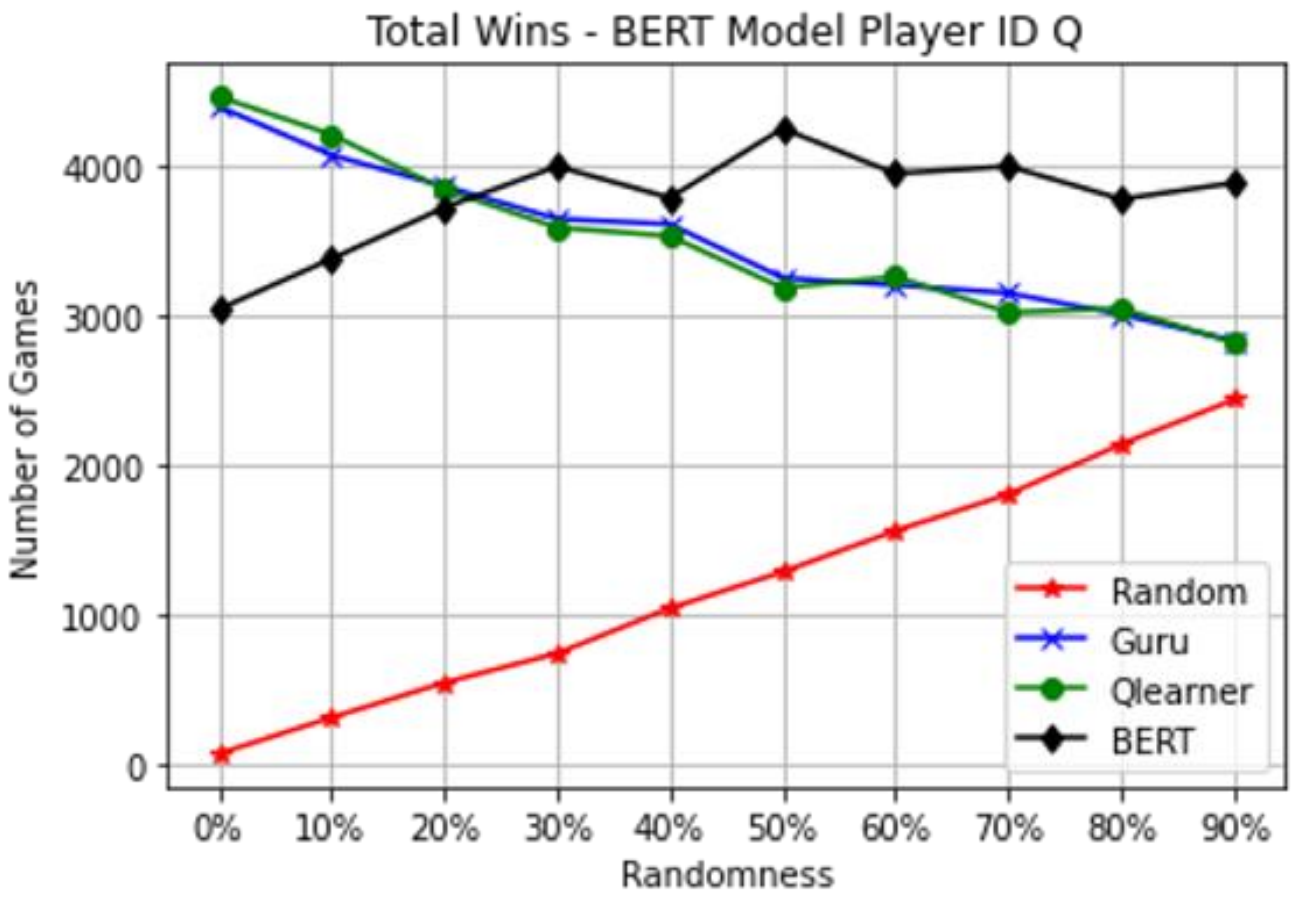

Figure 9. Nim Player ID Q. The BERT model inferenced and played with the Q learner (Q) ID token being specified.

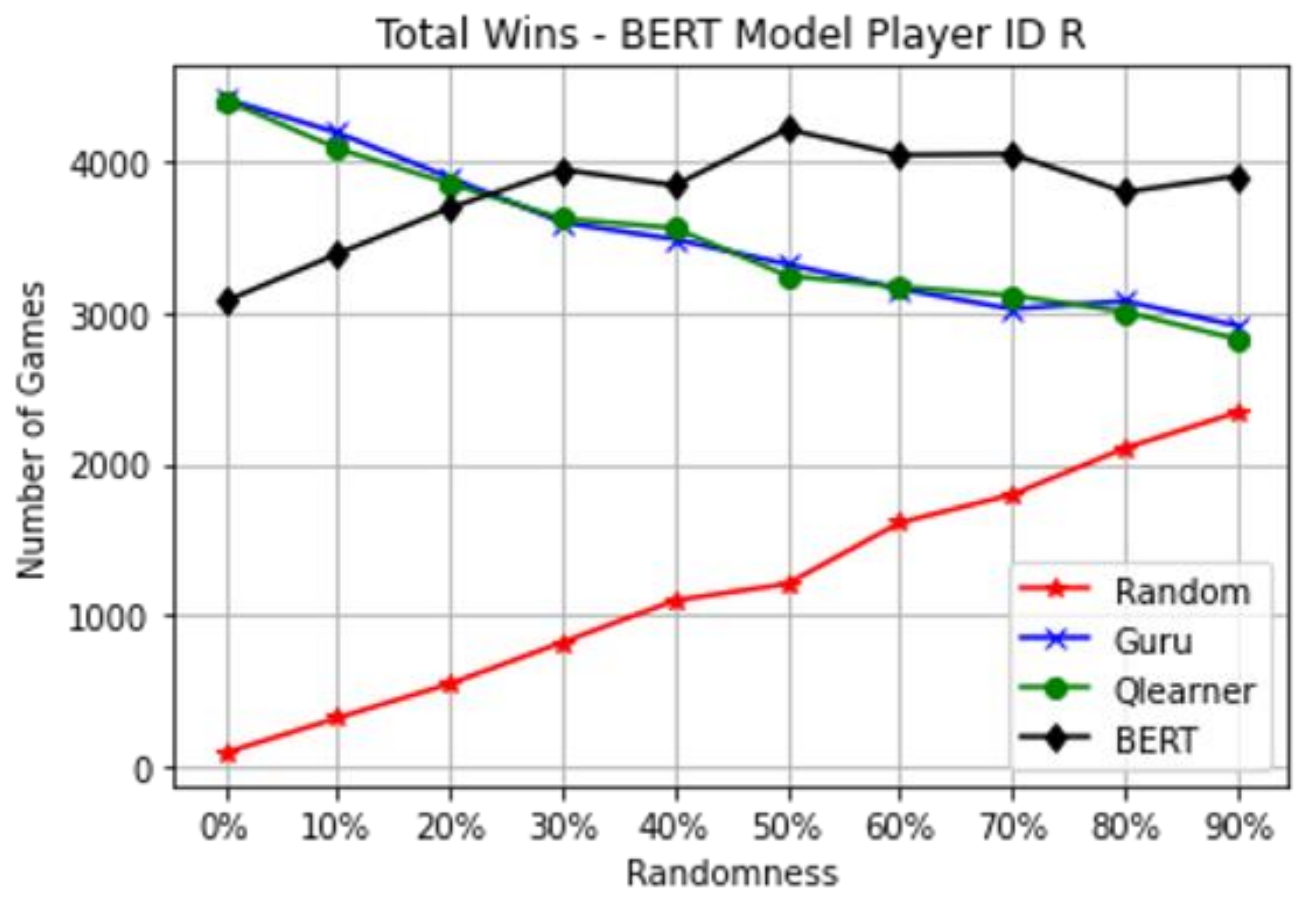

Figure 10. Nim Player ID R. The BERT model inferenced and played with the random agent (R) ID token being specified.

The BERT model consistently beats the other agents as the level of randomness increases in the dataset. This supports our original hypothesis because it shows evidence that the BERT model can identify the strongest signal (Guru and Q-learner) despite the random noise. This is especially evident in the 90% index of the results. Despite learning from a dataset where the players were only making one out of every ten of their moves, the model performed better than them. This is shown in Figure 7, Figure 9, and Figure 10 where at randomness threshold of 30% the BERT model outperforms all the agents. This also held true up till 100% random. The model did not perform well however with a win/loss indicator system (graphed in Figure 8). In fact, the model somewhat follows the same performance trend as the agents.

The process of playing as one player or the other is defined within the state space of the text sequence. This is because the text sequence for a Nim sequence is generalized as the game space followed by an indicator token, and the corresponding move. For example, the sequence a10/b10/c10 G – [MASK] indicates that this should be a Guru agent move and a sequence such as a10/b0/c0 W – [MASK] indicates this is a winning move. These types of indicators are encoded into the dataset, and the transformer model learned these patterns.

The role of these indicators in the performance of the model is interesting. As it shows that one could perform additionally postprocessing on the dataset to add more attributes from which the model could learn from.

## 3.2. Results Chess

Only one BERT word piece tokenizer was trained on the Chess dataset. Its total vocabulary size is 16,000 tokens so it is not possible to show here like the Nim vocabularies.

The chess dataset was created with the first three moves (for each max level chess engine) per game. This is important to keep in mind as the BERT model seemed to perform well given that it had a very small subset of all possible chess moves.

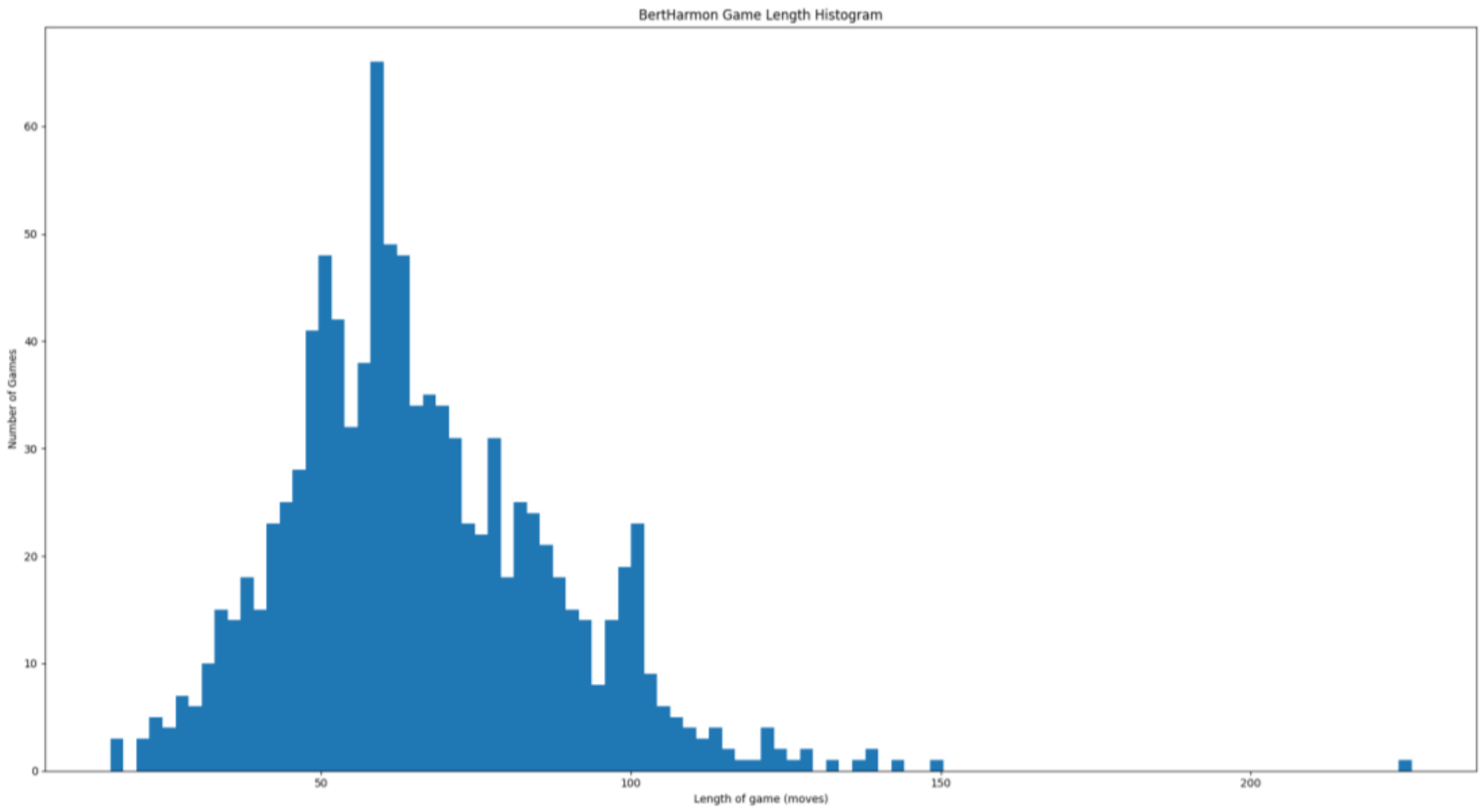

Figure 11. BERT Chess Game Length Distribution Versus Grandmaster Chess Engine

The BERT Chess model was not proficient enough to win chess games, so instead we show how well it did at playing chess against a grandmaster level chess player (Stockfish at max ELO). Additionally, it took around 3 days to train. Since the choice of a move given a board space is an open-ended answer, the model could technically answer with any text that it had in its vocabulary. We consider this as a feature of the model that it was given the option of answering in an incorrect format. As a result, we benchmarked its accuracy in terms of giving valid chess moves (shown in Figure 12). Given that it was only aware of the opening states of chess, it is impressive that at 35 moves into a game (35 moves for each player) it has an accuracy of 75%. When the model got a move wrong, we substituted a stockfish move in its place, and kept the game going.

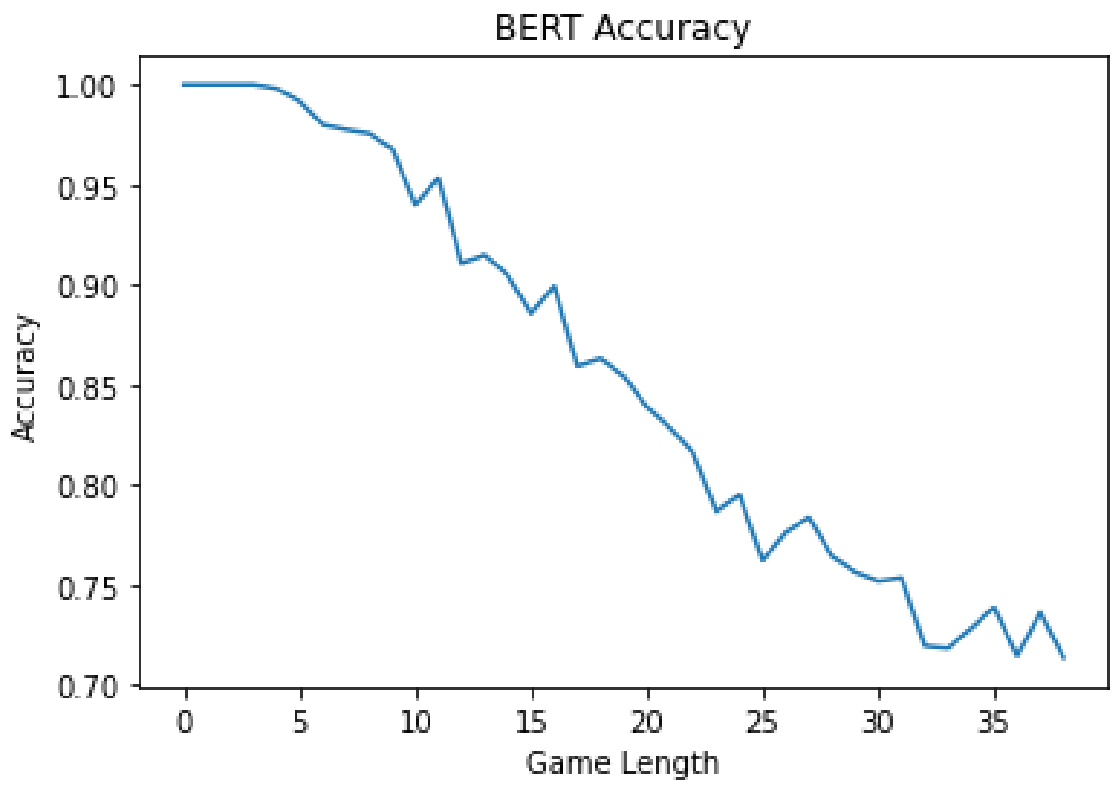

Figure 12. BERT Accuracy for Choosing Valid Chess Moves

In addition to measuring accuracy, we also measured the game length endurance of the model to understand how long it could play against a grandmaster stockfish engine till it lost. The results are graphed in Figure 11. Surprisingly BERT could survive for on average 32 moves (65 moves total for the game, ~32 per player), and at most we saw a game lasting for more than 200 moves.

# 4. CONCLUSIONS

In conclusion, we have shown that the BERT model is capable of learning both the games of Nim and Chess. We built text corpuses by representing game states and moves in a text sequence format. We have shown that the BERT transformer model is able to learn games in the context of very little information, in the presence of large quantities of noise, and in the presence of a large amount of data. The BERT model has been shown to learn the behaviors and patterns of primary game agents Guru, Q-learner, and stockfish such that the model can emulate their actions.

The results of our research should encourage BERT and other transformer models to be used as few-shot learners in situations where data is expensive to gather, difficult to clean, and in very high dimensional learning environments.

Transformer language models can represent input sequences efficiently through various autoencoding steps such as BERT, ALBERT, RoBERTa, ELECTRA, etc. Exploring the performance of these language models can help improving chess language models in this study. Second, these models can be used to cluster chess opening positions in order to compare and contrast to the ECO chess openings taxonomy. Future work will explore the clustering of chess positions to build taxonomy of openings, middle game positions and end game positions. This approach is analogous to text summarization where BERT approaches are known to be successfully applied. Third, future work will investigate player attribution in chess by analyzing various master games in chess databases as certain player styles are known to exist, such as Karpov likes closed and slow games, as Kasparov and Tal like open and sharp games.

Additionally, by representing a game space in our text sequence format, there are several interesting use cases with BERT such as authorship attribution, author playstyle and game space deduction. Given a dataset of games played by grandmasters, one could train this model and assess the probability that a given move has been made by Kasparov or another grandmaster by solving for the author/agent token instead of a move token. Additionally, if we solve for the move token then one could identify how a specific grandmaster would play given the board state. These two use cases allow for someone to prepare against a specific opponent. Given there are three spaces of the sequence, the last portion to solve for is the game space, and interestingly one could solve for the game space to suggest what is the most likely game space to precede this move for this player. All these use cases generalize for practical real-world problems that can be conceptualized into a state-independent text corpus such as predicting consumer behavior.

## ACKNOWLEDGEMENTS

Special thanks to Booz Allen Hamilton, Johns Hopkins University, as well as our friends and family for their support.

## AUTHORS

**Michael DeLeo** is an engineer and researcher. He currently works at Booz Allen Hamilton as a Machine Learning Engineer. He graduated from Penn State with a BS in Computer Engineering (minors in Math and Computer Science). He is also currently studying for his masters in Artificial Intelligence at Johns Hopkins University where he is performing research on NLP. Email ID: mdeleo2@jhu.edu

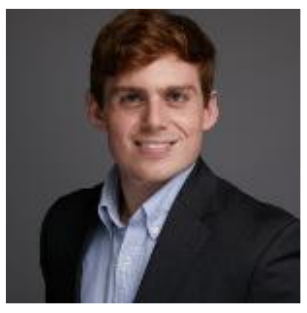

**Erhan Guven** is a faculty member at JHU WSE. He also works at JHU Applied Physics Lab as a data scientist and researcher. He received the M.Sc. and Ph.D. degrees from George Washington University. His research includes Machine Learning applications in speech, text, and disease data. He is also active in cybersecurity research, graph analytics, and optimization. Email ID: eguven2@jhu.edu

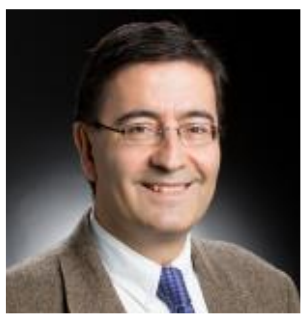